# Manipulability maximization in constrained inverse kinematics of surgical robots

*Jacinto Colan[1*], Ana Davila[2], Yasuhisa Hasegawa[1]*

[1]Department of Micro-Nano Mechanical Science and Engineering, Nagoya University, Aichi, Nagoya, Japan
[2]Institutes of Innovation for Future Society, Nagoya University, Aichi, Nagoya, Japan

*Corresponding Author: colan@robo.mein.nagoya-u.ac.jp*

*Abstract* - In robot-assisted minimally invasive surgery (RMIS), inverse kinematics (IK) must satisfy a remote center of motion (RCM) constraint to prevent tissue damage at the incision point. However, most of existing IK methods do not account for the trade-offs between the RCM constraint and other objectives such as joint limits, task performance and manipulability optimization. This paper presents a novel method for manipulability maximization in constrained IK of surgical robots, which optimizes the robot's dexterity while respecting the RCM constraint and joint limits. Our method uses a hierarchical quadratic programming (HQP) framework that solves a series of quadratic programs with different priority levels. We evaluate our method in simulation on a 6D path tracking task for constrained and unconstrained IK scenarios for redundant kinematic chains. Our results show that our method enhances the manipulability index for all cases, with an important increase of more than 100% when a large number of degrees of freedom are available. The average computation time for solving the IK problems was under 1ms, making it suitable for real-time robot control. Our method offers a novel and effective solution to the constrained IK problem in RMIS applications.

## I. INTRODUCTION

Minimally invasive surgery (MIS) is a surgical technique that reduces trauma and speeds up recovery by using thin instruments through small incisions to access the patient's body. However, MIS also poses several challenges for the surgeon, such as limited dexterity of conventional tools, poor depth perception and lack of haptic feedback [1]. Robot-assisted minimally invasive surgery (RMIS) overcomes some of these difficulties and offers benefits for both patients and surgeons, such as lower trauma, blood loss, infection risk, and recovery time. Recent advances in RMIS technology have enabled the creation of high-dexterous surgical tools [2], autonomous surgical tasks [3] and surgical flow analysis [4]. However, RMIS still faces significant issues in motion control and precision of the surgical robot, especially when dealing with complex anatomy and constrained workspace. One of the main problems in RMIS is inverse kinematics (IK), which determines a set of joint values to achieve a desired position and orientation of the end-effector. IK is essential for attaining accurate and smooth motion of the surgical tool while avoiding collisions and singularities.

However, most existing IK methods have limitations in terms of accuracy, efficiency, and robustness [5]. Some methods rely on analytical solutions that are fast but only applicable to simple robot structures and do not handle constraints [6]. Other methods use numerical solutions that are more general but computationally expensive and susceptible to local minima [7]. Moreover, most methods do not account for the manipulability of the robot, which is a common measure of its dexterity and ability to avoid singularities and represent an important problem in RMIS because it influences the performance and safety of the surgery [8].

In this paper, we propose a novel method for manipulability maximization in constrained inverse kinematics of surgical robots. Our method aims to optimize the robot's dexterity while satisfying the remote center of motion (RCM) constraint and joint limits. The RCM constraint ensures that the surgical tool rotates around a fixed point at the incision site to prevent tissue damage, while the joint limits constraint prevents the robot from reaching physically impossible configurations that could cause damage or instability. The constraints are handle by a hierarchical quadratic programming (HQP) framework that solves a sequence of quadratic programs with different priority levels [9]. The manipulability maximization problem is integrated into the hierarchical optimization framework as a least square problem, and a linearization representation enables real-time performance for robot control.

## II. RELATED WORKS

### A. Constrained inverse kinematics for RMIS

Constrained inverse kinematics is commonly found in robotic applications that perform multiple tasks at the same time. Various types of constraints have been integrated into the inverse kinematic problem, such as avoiding joint limits [10], avoiding obstacles [11], maintaining robot posture [12], and maximizing manipulability [13]. Kinematic redundancy is often used to satisfy these additional constraints [14].

In RMIS, the RCM restriction is an essential constraint to ensure safety of the patient. Typically, the RCM constraint is handle with passive mechanisms or programmable RCMs. Passive RCM mechanisms are usually fixed to the robot structure reducing adaptability. Programmable RCM, on the other hand, provide flexibility with software-based RCM that can dynamically change according to the surgery requirements. Multiple formulations of programmable RCMs have been proposed, including null-space projection [15], augmented Jacobian [16], trajectory generation [17], optimization-based [18] and concurrent approaches [19].

The constrained problem has been commonly addressed with prioritized strategies. Null-space projection is the most common approach to handle multiple constraints with different priorities [20]. However, analytical implementations can found limitations in handle inequality constraints, such as joint limits, and is susceptible to local minima. Optimization based, on the other hand, can include inequalities into the optimization problem, and handle task prioritization by assigning weights to each task according to its relative importance. It, however, cannot ensure strict prioritization. Hierarchical Quadratic Programming (HQP) solves hierarchically multiple tasks as QP problems based on their priority [9]. HQP is a powerful technique that can handle multiple objectives and constraints in a consistent and efficient way.

### B. Manipulability maximization

The manipulability index is a common measure of the dexterity and ability of a robot to avoid singularities [21]. The manipulability represents the ability of the manipulator to generate linear/angular velocities at each joint and the end-effector for a given configuration. For redundant robot manipulators, the higher number of DOFs can be utilized to enhance dexterity for complex tasks by maximizing its manipulability [22].

Several methods have been proposed to maximize manipulability in motion planning. A Kalman filter has been proposed for manipulability optimization of a 6-DOF manipulator, but susceptible to local minima [23]. Bayle et al. [24] proposed the use of a project gradient method. In [25], the optimization problem is formulated as a constrained quadratic programming problem. In [26], a continuous-time trajectory optimization that leverages a Gaussian process representation of the trajectory and a maximum a posteriori estimator is used to maximize manipulability for a 6-DOF manipulator. Leoro et al. [27] proposed the manipulability measure that combines the normalized manipulability of the base and arm for nonholonomic mobile manipulators. In RMIS, manipulability maximization represents an important problem to ensure safety during a surgical task. However, additional constraints such as RCM, need to be considered simultaneously, and this aspect has been scarcely addressed in previous works. For instance, Su et al. [8] devised a neural network based method for maximizing the manipulability of redundant manipulators.

### III. METHODOLOGY

#### A. Multi-task Hierarchical optimization

Surgical robotic applications require to accomplish several tasks simultaneously. Optimization frameworks for constrained inverse kinematics problems have proven useful by integrating inequality constraints in the optimization process and avoiding getting stuck in local minimas. A hierarchical optimization approach allows for simultaneous execution of multiple tasks, each with different levels of priority. We follow a Hierarchical Quadratic Programming approach (HQP), in which each priority-level tasks are solved as a QP problem, and strict prioritization avoids lower-priority task affecting higher-priority tasks [9]. Between tasks in the same priority-level, weights can be assigned to each task for soft prioritization.

In general, given $n$ tasks, each with weights $K_{t_i}$ representing their priority, the soft constrained inverse kinematics problem for a kinematic chain with $n_q$ joints can be defined as:

$$\min_{\dot{q},w} \sum_{i=1}^{n} \frac{K_{t_i}}{2} \|J_{t_i}\dot{q} - K_{r_i}e_{t_i}\|^2 + \frac{K_d}{2}\|\dot{q}\|^2 + \frac{K_w}{2}\|w\|^2$$
s.t. (1)
$$C_p\dot{q} - d_p \leq w$$

where $J_{t_i}$, $e_{t_i}$ and $K_{r_i}$ represent the Jacobian, residual and residual gain for task $i \in [1, n]$. The joint configuration is represented by $q \in \mathbb{R}^{n_q}$. The slack variable $w$ is used for the inequalities constraints and the square norm of the joint velocities given by $\|\dot{q}\|^2$ works as regularization term. $C_p$ and $d_p$ are the matrix and vector representing the inequality constraints. The $K_d$ and $K_w$ are positive constant gains for the damping and slack components.

The equation in (1) can be represented as a Quadratic Programming (QP) problem given by [28]

$$\frac{1}{2}x^T Q x + p^T x. \quad (2)$$

Where $x = [\dot{q} \ w]$ represents the optimization variable. The positive matrix $Q$ and vector $p$ are given by

$$Q = \overline{A_{p_1}}^T \overline{A_{p_1}} \qquad p = -\overline{A_{p_1}}^T \overline{b_{p_1}} \quad (3)$$

with

$$A_{p_1} = \begin{bmatrix} K_{t_1}^{1/2} A_1 \\ \vdots \\ K_{t_n}^{1/2} A_n \\ K_d^{1/2} I \end{bmatrix} \quad b_{p_1} = \begin{bmatrix} K_{t_1}^{1/2} b_1 \\ \vdots \\ K_{t_n}^{1/2} b_n \\ 0 \end{bmatrix} \quad (4)$$

$$\overline{A_{p_1}} = \begin{bmatrix} A_{p_1} & 0 \\ 0 & K_w^{1/2} I \end{bmatrix} \qquad \overline{b_{p_1}} = \begin{bmatrix} b_{p_1} \\ 0 \end{bmatrix}$$

While the previous representation allows for soft prioritization between tasks, some task need to be strictly ensured for safety considerations (e.g. RCM constraints). To ensure hard-prioritization, the lower priority tasks are solved in the null space of the higher priority level. The QP problem for the lower priority tasks can then be defined by

$$\min_{\dot{q},w} \sum_{i=1}^{T} \frac{K_{t_i}}{2} \|J_{t_i}(N_{p-1}\dot{q} + \bar{q}_{p-1}^*) - K_{r_i}e_{t_i}\|^2 + \frac{K_d}{2}\|\dot{q}\|^2$$
$$+ \frac{K_w}{2}\|w\|^2$$
s.t. (5)
$$C_p(N_{p-1}\dot{x} + \bar{x}_{p-1}^*) - d_p \leq w$$
$$C_{p-1}(N_{p-1}\dot{x} + \bar{x}_{p-1}^*) - d_{p-1} \leq w_{p-1}^*$$
$$\vdots$$
$$C_1(N_{p-1}\dot{x} + \bar{x}_{p-1}^*) - d_1 \leq w_1^*$$

where $N_{p-1}$ represents the null space projector for the higher priority level. Matrix $Q$ and vector $p$ are given by (3), with

$$A_{p_1} = \begin{bmatrix} K_{t_1}^{1/2} A_1 \\ \vdots \\ K_{t_n}^{1/2} A_n \\ K_d^{1/2} I \end{bmatrix} \quad b_{p_1} = \begin{bmatrix} -K_{t_1}^{\frac{1}{2}} (A_1) \\ \vdots \\ K_{t_n}^{1/2} b_n \\ 0 \end{bmatrix} \quad (6)$$

$$\overline{A_{p_1}} = \begin{bmatrix} A_{p_1} & 0 \\ 0 & K_w^{1/2} I \end{bmatrix} \quad \overline{b_{p_1}} = \begin{bmatrix} b_{p_1} \\ 0 \end{bmatrix}$$

The optimal solution $\bar{x}_p^*$ is then given by

$$\bar{x}_p^* = N_{p-1}\bar{x}_{p-1}^*. \tag{7}$$

where $\bar{x}_{p-1}^*$ represent the optimal solutions for the previous priority level.

### B. Manipulability maximization

The Manipulability Index is based on the Jacobian Matrix and defined by [21]

$$m(q) = \sqrt{det(JJ^T)} \tag{8}$$

Maximization of the manipulability index can be formulated as

$$\min_{\dot{q}} -m(q) \tag{9}$$

However, this represents a highly nonlinear optimization problem. This can be addressed by linearization from the Taylor expansion of (1) given as

$$m(q + \Delta q) = m(q) + \Delta t \, \nabla m^T \dot{q} + \tfrac{1}{2}\Delta t^2 \dot{q}^T H_m \dot{q} \tag{10}$$

Where $\Delta t$ is the sampling time, and $\nabla m$ and $H_m$ are the gradient vector and Hessian matrix of $m$ respecitvely. The maximization problem defined in (9) can now be represented as:

$$\min_{\dot{q}} -\left(\Delta t \, \nabla m^T \dot{q} + \tfrac{1}{2}\Delta t^2 \dot{q}^T H_m \dot{q}\right) \tag{11}$$

To avoid computing $H_m$, which is computationally expensive for real-time applications, the following equivalent optimization problem is used [13]

$$\min_{\dot{q}} \; \tfrac{1}{2}\Delta t^2 \dot{q}^T \nabla m_k \nabla m_k^T \dot{q} + m\Delta t \, \nabla m^T \dot{q} \tag{12}$$

which represents a least square problem defined by

$$\min_{\dot{q}} \; \|\Delta t \nabla m_k^T \dot{q} - m_k\|^2 \tag{13}$$

For the online gradient computation, we followed a numerical approach as proposed in [13].

### C. End-effector pose control

The end-effector pose control problem is given by

$$\min_{\dot{q}} \; \|J_{ee}\dot{q} - e_{ee}\|^2, \tag{14}$$

Where $J_{ee}$ represents the end-effector Jacobian and $e_{ee}$ is the end-effector task residual defined by

$$e_{ee} = \log_6({}_B^B T_{ee_{des}} {}_B^B T_{ee_{act}}^T), \tag{14}$$

where the $\log_6: SE(3) \to se(3)$ maps the pose from the Lie group $SE(3)$ to twists in the $se(3)$. The homogenous transformations ${}_B^B T_{ee_{des}}$ and ${}_B^B T_{ee_{act}}$ represents the desired and actual end-effector pose.

### C. RCM constraint

The RCM constrained can be represented as a least square problem [19] given by

$$\min_{\dot{q}} \; \|J_{rcm}\dot{q} - e_{rcm}\|^2, \tag{14}$$

where the RCM task Jacobian is defined by

$$J_{rcm} = \tfrac{\delta p_{rcm}}{\delta q} = (I - \hat{p}_s \hat{p}_s^T) J_{pre} + (\hat{p}_s p_r^T + p_r^T \hat{p}_s I) \tfrac{\delta \widehat{p_s}}{\delta q} \tag{15}$$

with

$$\tfrac{\delta \widehat{p_s}}{\delta q} = \tfrac{1}{\|p_s\|}(I - \hat{p}_s \hat{p}_s^T)(J_{pre} - J_{post}), \tag{16}$$

and the RCM residual is given by

$$e_{rcm} = \|p_{trocar} - p_{rcm}\|. \tag{17}$$

### D. Joint limits

The joint limits are represented by the inequality

$$\tfrac{q^- - q}{\delta t} \leq \dot{q} \leq \tfrac{q^+ - q}{\delta t} \tag{17}$$

where $q^-$ and $q^+$ represent the lower and upper joint limits, while $\delta t$ is the cycle time. The slack variable $w = [w^+ \; w^-]$ is used to integrate the joint limits into the optimization problem.

$$\dot{q} - \tfrac{q^+ - q}{\delta t} \leq w^+ \tag{18}$$

$$-\dot{q} + \tfrac{q^- - q}{\delta t} \leq w^-$$

The matrix $C_1$ and $d_1$ can then be represented as

$$d_1 = \begin{bmatrix} \tfrac{q^+ - q}{\delta t} \\ -\tfrac{q^- - q}{\delta t} \end{bmatrix} \quad C_1 = \begin{bmatrix} I \\ -I \end{bmatrix} \tag{19}$$

### E. Constrained inverse kinematics

The inverse kinematics problem is formulated for three tasks in two priority-levels. The higher priority level contains the RCM task, and is formulated as in (1) with $n = 1$, and

$$A_1 = J_{rcm} \quad b_1 = e_{rcm} \tag{19}$$

The lower priority level includes the end-effector pose control and the manipulability maximization tasks ($n = 2$), each assigned with a weight representing its soft priority. The optimization problem is formulated as (5) with

$$A_2 = J_{ee} \quad b_2 = e_{ee} \tag{20}$$
$$A_3 = \Delta t \nabla m_k^T \quad b_3 = m_k \tag{21}$$

## IV. EXPERIMENTAL VALIDATION

We evaluated the performance of the proposed method in a simulation environment developed in Coppeliasim [29]. Two redundant kinematic chains designed for RMIS applications were studied:

- KC-1: A 7-DOF manipulator with a 3-DOF robotic surgical tool [2], shown in Fig. 1.
- KC-2: A 7-DOF manipulator with a 5-DOF robotic surgical tool, shown in Fig. 2.

The IK solver was implemented in C++ on a Linux Ubuntu 20.04 workstation with an Intel Core i9-11900 processor and 64 GB of RAM. The implementation was designed to be compatible with the robotics operating system (ROS) framework. For kinematic computations, transformations, and kinematic chain parsing, we employed the Pinocchio library (v. 2.6.10) [30]. We used CASadi (v. 3.5.5) [31] as a backend for nonlinear and HQP solvers with OSQP (v. 0.5.0) [32] for quadratic programming in the HQP solver, both with warm start enabled.

The target task consists of tracking a 6D path. Two scenarios were evaluated: a constrained case with an RCM restriction as high priority, and an unconstrained case without RCM.

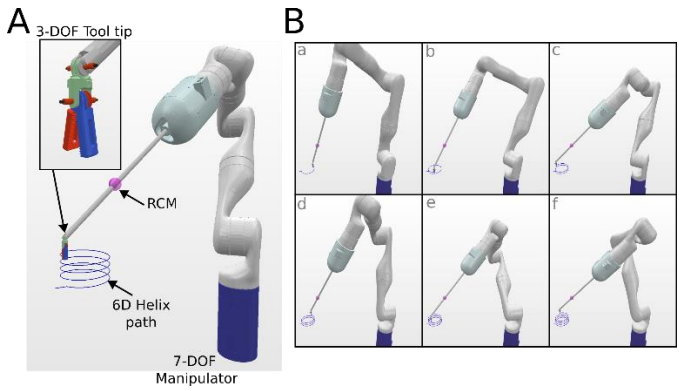

Fig. 1 Kinematic chain KC-1. **A.** The surgical robot consists of a 7-DOF manipulator and a 3-DOF surgical tool. The helix path followed is shown in blue and the RCM in magenta. The inset zooms in the tool tip. **B.** Snapshots in sequence of the robot performance tracking the helix path.

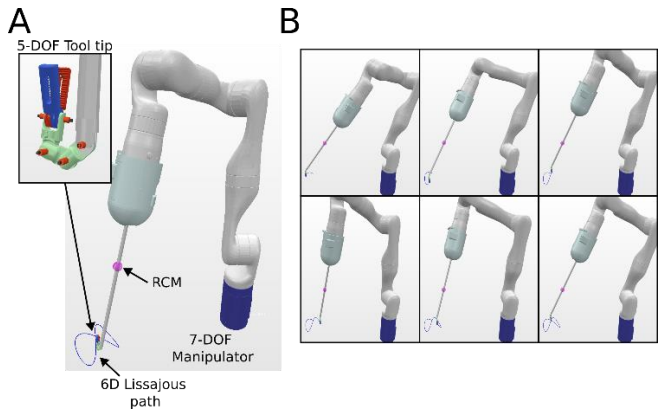

Fig. 2 Kinematic chain KC-2. **A.** The surgical robot consists of a 7-DOF manipulator and a 5-DOF surgical tool. The Lissajous path followed is shown in blue and the RCM in magenta. The inset zooms in the tool tip. **B.** Snapshots in sequence of the robot performance tracking the helix path.

### A. RCM –constrained scenario

The parameters used for the RCM-constrained path tracking are described in Table 1.

TABLE I
PARAMETERS FOR THE RCM-CONSTRAINED TASK

| $K_{t_1}$ | $K_{t_2}$ | $K_{t_3}$ | $K_{r_1}$ | $K_{r_2}$ | $K_{r_3}$ | $K_{d_1}$ | $K_{d_2}$ | $K_{w_1}$ | $K_{w_2}$ |
|---|---|---|---|---|---|---|---|---|---|
| 1.0 | 1.0 | 0.01* | 1.0 | 1.0 | 1e-3 | 1e-5 | 1e-9 | 1e-5 | 1e-5 |

*For KC-2, $K_{t_3} = 0.001$

The tracking target is a 6-DOF helix of 7cm diameter described by

$$r(t) = \begin{bmatrix} x_0 \\ y_0 \\ z_0 \end{bmatrix} + \begin{bmatrix} A\cos 2\pi t \\ A\sin 2\pi t \\ 0.01t \end{bmatrix} \quad (22)$$

With the target orientation fixed to $R_d = \begin{bmatrix} 1 & 0 & 0 \\ 0 & 0 & 1 \\ 0 & -1 & 0 \end{bmatrix}$, which can be observes as the tip pointing downwards. We compared the average manipulability index, the maximum manipulability index, the average RCM error and the average pose error. The results are summarized in Table II.

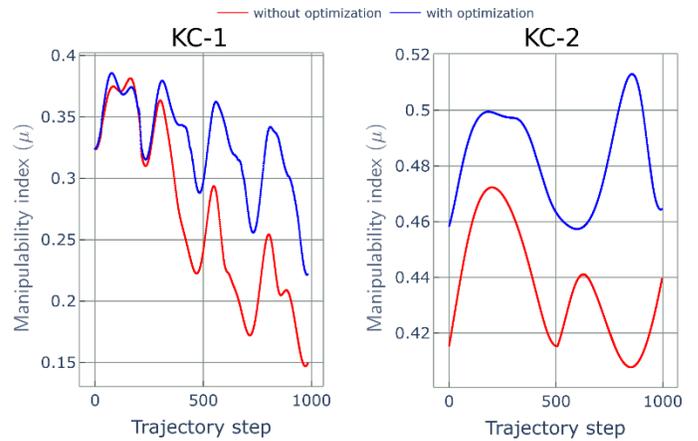

Fig. 3 Evolution of the manipulability index with respect to the trajectory step for both kinematic chains (KC-1 and KC-2) in an RCM-constrained scenario. The proposed method (represented in blue) shows a higher manipulability for almost all trajectory steps.

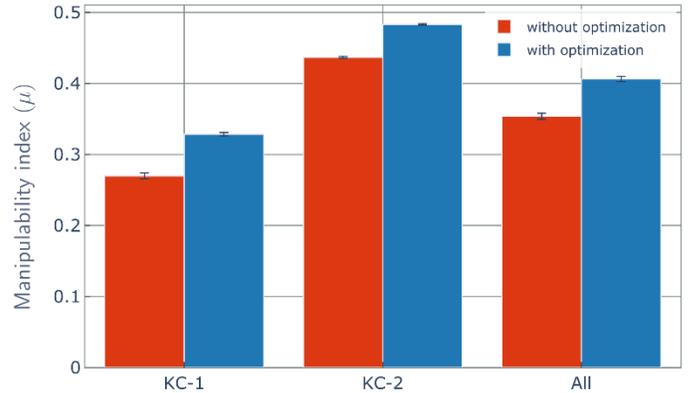

Fig. 4 Comparison of the average manipulability index for both kinematic chains (KC-1 and KC-2) in an RCM-constrained scenario. The proposed method (represented in blue) shows an increment of the manipulability index compared to without optimization (represented in red) of about 15%.

TABLE II
RESULTS FOR THE CONSTRAINED TASK

| | KC-1 | | KC-2 | |
|---|---|---|---|---|
| | Without optimization | With optimization | Without optimization | With optimization |
| Avg. Manipulability | 0.270 | **0.329** | 0.437 | **0.483** |
| Max. Manipulability | 0.381 | **0.386** | 0.472 | **0.512** |
| Avg. RCM error ($e_{rcm}$) (mm) | 0.0056 | 0.0061 | 0.0049 | 0.0070 |
| Avg. EE pose error ($e_{ee}$) | 6.6x10$^{-6}$ | 11x10$^{-6}$ | 0.7x10$^{-6}$ | 0.6x10$^{-6}$ |

Figure 3 shows the manipulability index for both, without and with manipulability optimization with respect to the trajectory step. It can be observed that for both cases the proposed method increases the manipulability index. Figure 4 depicts the average manipulability index for each kinematic chain. For the kinematic chain KC-1, the manipulability index incremented in 21%, while for the KC-2 case was about 11%.

For KC-2, which have more additional degrees of freedom available, the maximum manipulability index increases in 8%. The average RCM error and EE error remained within the desired range of under 0.0001.

### B. Unconstrained scenario

For the unconstrained case, the 6D path remains the same, but the RCM constraint was removed. The parameters used for the unconstrained path tracking are the same as described in Table I. The target path is a 6-DOF Lissajous curve defined by

$$r(t) = \begin{bmatrix} x_0 \\ y_0 \\ z_0 \end{bmatrix} + \begin{bmatrix} A \sin t \\ B \sin(2t + \pi) \\ C(\cos 2t - 1) \end{bmatrix} \quad (23)$$

The orientation is fixed to $R_d = \begin{bmatrix} 1 & 0 & 0 \\ 0 & 0 & 1 \\ 0 & -1 & 0 \end{bmatrix}$, which can be observed in the figure as pointing upwards. The results are summarized in Table III.

TABLE III
RESULTS FOR THE UNCONSTRAINED TASK

|  | KC-1 | | KC-2 | |
| --- | --- | --- | --- | --- |
|  | Without optimization | With optimization | Without optimization | With optimization |
| Avg. Manipulability | 0.383 | **0.406** | 0.334 | **0.824** |
| Max. Manipulability | 0.416 | **0.444** | 0.376 | **0.910** |
| Avg. EE pose error ($e_{ee}$) | $1.1 \times 10^{-6}$ | $1.5 \times 10^{-6}$ | $0.4 \times 10^{-6}$ | $1.9 \times 10^{-6}$ |

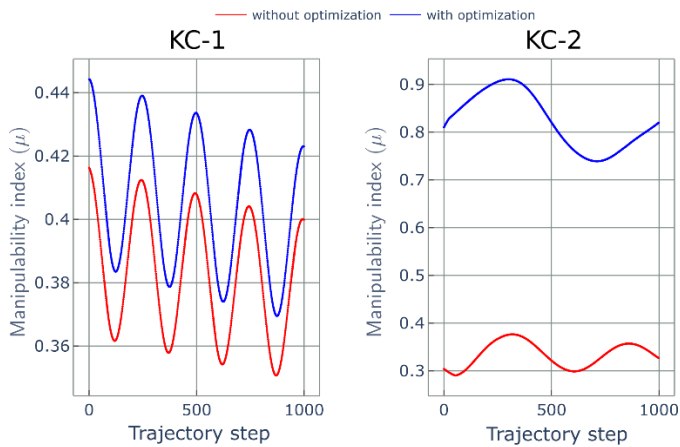

Fig. 5 Evolution of the manipulability index with respect to the trajectory step for both kinematic chains (KC-1 and KC-2) in an RCM-constrained scenario. The proposed method (represented in blue) shows a higher manipulability for all trajectory steps with larger improvement for KC-2.

Figure 5 shows the manipulability index with respect to the trajectory step. It can be observed that for both cases the proposed method increases the manipulability index, with a larger range for the KC-2. Figure 6 depicts the average manipulability index for each kinematic chain. For the kinematic chain KC-1, the manipulability index incremented in 6%, while for the KC-2 case was about 146%. For KC-2, which have more additional degrees of freedom available, the maximum manipulability index increases in 142%. The average RCM error and EE error remained within the desired range of under 0.0001.

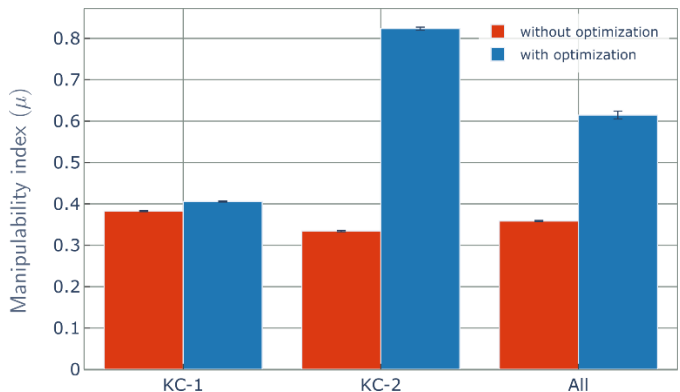

Fig.6 Comparison of the average manipulability index for both kinematic chains (KC-1 and KC-2) in an RCM-constrained scenario. The proposed method (represented in blue) shows an increment of the manipulability index compared to without optimization (represented in red) of about 60%.

### C. Computation time performance

For real-time robot control, computation time for IK solving is critical. Figure 7 show the runtime for the IK solving with the proposed method. The constrained case required a mean computation time of less than 1ms, while the unconstrained case exhibitted computation times of less than 0.3 ms. The fast solving time demonstrates the capability of the proposed method for real-time robot control implementations.

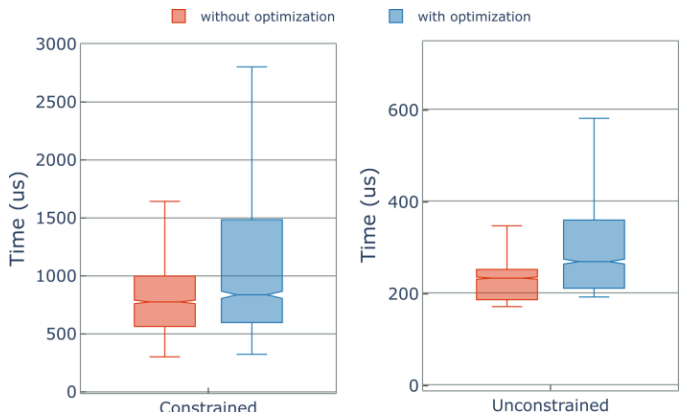

Fig. 7 Computation time for the constrained IK solving. When manipulability is optimized (represented in blue), there is an increment in the mean computation time of about 10% compared to without optimization (represented in red).

## V. CONCLUSIONS

Constrained inverse kinematics in surgical robots is a challenging problem requiring to achieve multiple task simultaneously. Manipulability has been widely used to improve performance of robotic systems in terms of singularity avoidance and dexterity. In this work, we integrate a manipulability maximization optimization problem into a hierarchical framework for constrained inverse kinematics that includes end-effector pose control, RCM constraint and joint limits. The results in simulation show that the proposed approach increases the manipulability index in constrained and unconstrained scenarios, with better performance with large degrees of freedom available. The mean IK solving time was under 1ms, proving to be effective for real-time robot control. Future work will focus on the integration of additional constraints such as collision avoidance.

ACKNOWLEDGMENT

This work was supported in part by the Japan Science and Technology Agency (JST) CREST under Grant JPMJCR20D5, and in part by the Japan Society for the Promotion of Science (JSPS) Grants-in-Aid for Scientific Research (KAKENHI) under Grant 22K14221.